# Exponentiated Gradient LINUCB for Contextual Multi-Armed Bandits

Djallel Bouneffouf

Department of Computer Science, Télécom SudParis, UMR CNRS Samovar, 91011 Evry Cedex, France

`{Djallel.Bouneffouf}@it-sudparis.eu`

**Abstract.** We present Exponentiated Gradient LINUCB, an algorithm for contextual multi-armed bandits. This algorithm uses Exponentiated Gradient to find the optimal exploration of the LINUCB. Within a deliberately designed offline simulation framework we conduct evaluations with real online event log data. The experimental results demonstrate that our algorithm outperforms surveyed algorithms.

**Keywords:** machine learning; exploration/exploitation dilemma; artificial intelligence, LINUCB, exponentiated gradient.

## 1 Introduction

The bandit algorithm addresses then the need for balancing exploration/exploitation (exr/exp) tradeoff in different real life problem. A bandit algorithm B exploits its past experience to select actions that appear more frequently. Besides, these seemingly optimal actions may in fact be suboptimal, because of the imprecision in B's knowledge. In order to avoid this undesired case, B has to explore actions by choosing seemingly suboptimal actions so as to gather more information about them. Exploitation can decrease short-term user's satisfaction since some suboptimal actions may be chosen. However, obtaining information about the actions' average rewards (i.e., exploration) can refine B's estimate of the actions' rewards and in turn increases long-term user's satisfaction. Clearly, neither a purely exploring nor a purely exploiting algorithm works well, and a good tradeoff is needed. One interesting solution to the multi-armed bandit problem is the LINUCB strategy. This is an extension of the UCB algorithm to the parametric case [9]. It selects the document for which mean standard deviation is maximum.

We propose to improve this approach by adding a random exploration $\varepsilon$, also the $\varepsilon$ parameter controls essentially the exp/exr tradeoff between exploitation and exploration is computed using the exponentiated gradient .

One drawback of this algorithm is that it is difficult to decide in advance the optimal value. Instead, we introduce an algorithm named Gradien-LINUCB that achieves this goal by balancing adaptively the exp/exr tradeoff using gradient optimization.

The rest of the paper is organized as follows. Section 2 gives the key notions used throughout this paper. Section 3 reviews some related works. Section 4 describes the algorithms involved in the proposed approach. The experimental evaluation is illustrated in Section 5. The last section concludes the paper and points out possible directions for future work.

## 2     Key Notions

We formally define contextual bandit problems and bandit algorithms .
**Definition (Contextual bandit problem) .**
In a contextual bandits problem, there is a distribution P over $(x, r_1, ..., r_k)$, where x is context, $a \in \{1, ..., k\}$ is one of the k arms to be pulled, and $r_a \in [0,1]$ is the reward for arm a. The problem is a repeated game: on each round, a sample $(x, r_1, ..., r_k)$ is drawn from P, the context x is announced, and then for precisely one arm a chosen by the player, its reward $r_a$ is revealed.

**Definition (Contextual bandit algorithm) .**
A contextual bandits algorithm B determines an arm $a \in \{1, ..., k\}$ to pull at each time step t, based on the previous observation sequence $(x_1, a_1, r_a, 1), ..., (x_{t-1}, a_{t-1}, r_{a,t-1})$, and the current context $x_t$.).

## 3     Related Work

Very frequently used in reinforcement learning to study the exr/exp tradeoff, the multi-armed bandit problem was originally described by Robbins [7].
Few research works are dedicated to study the contextual bandit problem on recommender systems, where they consider the user's behavior as the context of the bandit problem. In [6], the authors extend the ε-greedy strategy by dynamically updating the ε exploration value. At each iteration, they run a sampling procedure to select a new ε from a finite set of candidates. The probabilities associated to the candidates are uniformly initialized and updated with the Exponentiated Gradient (EG) [5]. This updating rule increases the probability of a candidate $\varepsilon$ if it leads to a user's click.
In [9], authors model the recommendation as a contextual bandit problem. They propose an approach in which a learning algorithm sequentially selects documents to serve users based on their behavior. To maximize the total number of user's clicks, this work proposes LINUCB algorithm that is computationally efficient.

As shown above, both LINUCB and EG-greedy are interesting approach to solve the context bandit algorithm. Our intuition is that, combining these approaches lead to improve the result. This strategy uses the LINUCB and adds the EG for defining the optimal exploration (ε).

### 3.1 The LINUCB algorithm

Assuming that the expected reward of a document is linear in its features, the authors propose an approach in which a learning algorithm selects sequentially documents to serve users based on contextual information about the users and the documents.

```
Algorithm 2 The LINUCB algorithm
Input: α ∈ R+
for t=1 to T do
Observe features of all documents a ∈ At : xt;a ∈ Rd
for all a ∈ At do
if a is new then
Aa ← Id (d-dimensional identity matrix)
ba ← 0d*1 (d-dimensional zero vector)
end if
θa ← A⁻¹a * ba
pt;a ← θaT xt;a + √α xt;aT Aa⁻¹ xt;a
end for
Choose document at = argmaxa∈At pt;a with ties broken arbitrari-
ly, and observe a real-valued payoff rt
Aat ← Aat + xt;at xt;at
bat ← bat + rt xt,at
end for
```

In Alg. 1, $A_t$ is the set of documents $a$ at the iteration $t$, where $x_{a,t}$ is the feature vector of the a with d-dimension, θ is the unknown coefficient vector of the feature $x_{a,t}$ is a constant and $A_a = D_a^T D_a + I_d$. Da is a design matrix of dimension $m*d$ at trial $t$, whose rows correspond to m training inputs (e.g., m contexts that are observed previously for document a), and $b_a \in R^m$ is the corresponding response vector (e.g., the corresponding m click/no-click user feedback). Applying ridge regression to the training data $(D_a, c_a)$ gives an estimate of the coecients: $θ_a = (D_a^T D_a + I_d)^{-1} D_a^T c_a$, where $I_d$ is the d*d identity matrix.

### 3.2 The Gradien-LINUCB algorithm

To consider the random exploration on the LINUCB algorithm, the *Gradien-LINUCB* algorithm update the exploration value ε dynamically. In each iteration, they run a sampling procedure to select a new ε from a ε nite set of candidates.
The probabilities associated to the candidates are uniformly initialized and updated-with the Exponentiated Gradient (EG) [34]. This updating rule increases the probability of a candidate ε if it leads to a user's click.

```
Algorithm 2 The Gradien-LINUCB
Input: (ε1,…, εT) : candidate values for ε,
UC(d1,…,dn):documents
```

```
τ,β and k: parameters for EG N: number of iterations
Pk← 1 = T and wk ←1; k = 1,…,T
for i=1 to N do
Sample εd from Discrete (p1,…, pT )
```
$$d_i = \begin{cases} LINUCB_{UC}(CTR(d)) & if (q>\varepsilon_d) \\ Random(UC) & otherwise \end{cases}$$
```
make recommendation with di
Receive a click feedback ci from the user
wk ( wk exp(τ (ci I(k = d)+β)/pk ) , k = 1, ... , T
pk ← (1 - k)(wk/(∑Tj=1 wj) + k=/T); k = 1, ... , T
end for
```

In Alg. 2, p = (p$_1$, ... , p$_T$ ), where p$_i$ stands for the probability of using ε$_i$ for random exploration. These probabilities are initialized to be 1/T at the beginning, w = (w$_1$, ... , w$_T$ ) are a set of weights to keep track of the performance of each ε$_i$. Noting that the probability p is computed by normalizing w with smoothing, where I[z] is the indictor function and τ and β are a smoothing factors in weights updating. k is a regularization factor to handle singular w$_i$.

## 4  Experimental Evaluation

In order to empirically evaluate the performance of our approach, We now consider an online advertising application.

Given a user visiting a publisher page, the problem is to select the best advertisement for that user. A key element in this matching problem is the click-trough rate (CTR) estimation: what is the probability that a given ad will be clicked given some context (user, page visited)? Indeed, in a cost-per-click (CPC) campaign, the advertiser only pays when his ad gets clicked. is the reason why it is important to select ads with high CTRs. There is of course a fundamental exploration / exploitation dilemma here: in order to learn the CTR of an ad, it needs to be displayed, leading to a potential loss of short-term revenue.

More details on the data and on display advertising cab be found in (Agarwal et al., 2010).

To test the proposed *Gradien-LINUCB* algorithm, The testing step consists of evaluating the algorithms for each testing iteration using the average CTR. The average CTR for a particular iteration is the ratio between the total number of clicks and the total number of displays. Then, we calculate the average CTR over every 1000 iterations. The number of documents (N) returned by the recommender system for each iteration is 10 and we have run the simulation until the number of iterations reaches 10000, which is the number of iterations where all algorithms have converged.

   In the first experiment, in addition to a pure exploitation baseline, we have compared our algorithm to the algorithms described in the related work (Section 3): *EG-greedy*; *LINUCB*. In Fig. 2, the horizontal axis is the number of iterations and the vertical axis is the performance metric.

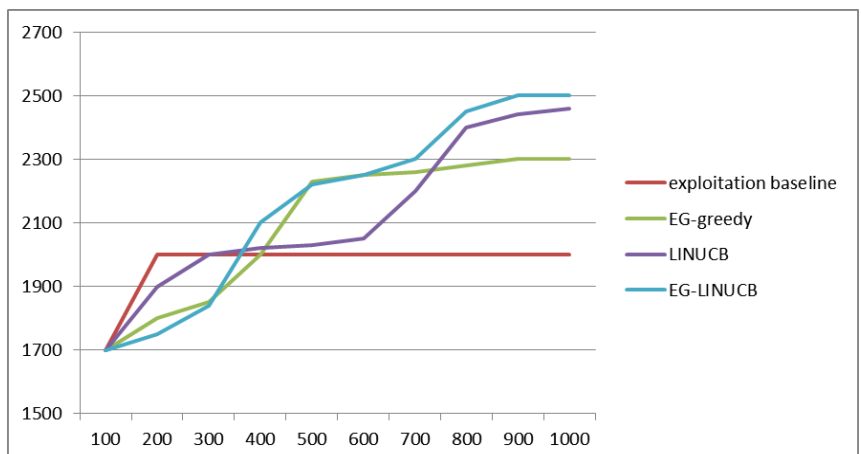

**Fig. 1.** Average CTR for exr/exp algorithms

We have several observations regarding the different exr/exp algorithms. For the *ε-decreasing* algorithm, the converged average CTR increases as the *ε* decreases (exploitation augments).

While the *EG* algorithm converges to a higher average CTR, its overall performance is not as good as *LinUCB*. Its average CTR is low at the early step because of more exploration, but does not converge faster. The *Gradien-LINUCB* algorithm effectively learns the optimal *ε*; it has the best convergence rate, increases the average CTR by a factor of 1,5 over the baseline and outperforms all other exr/exp algorithms. The improvement comes from a dynamic tradeoff between exr/exp, controlled by the EG estimation.

## 5      Conclusion

In this paper, we have presented an Exponentiated Gradient LINUCB algorithm for contextual multi-armed bandits. This algorithm uses Exponentiated Gradient to find the optimal exploration of the LINUCB.

The experimental results demonstrate that our algorithm performs better on average CTR in various configurations. In the future, we plan to evaluate the scalability of the algorithm on-board a mobile device and investigate other public benchmarks.